%% file: main.tex
\documentclass[10pt,twocolumn,letterpaper]{article}

\usepackage{verbatim}
\usepackage{booktabs}       
\usepackage{style}
\usepackage[pagenumbers]{cvpr} 
\usepackage{float}
\usepackage{multirow}
\usepackage{threeparttable}
\usepackage[accsupp]{axessibility} 

\input{preamble}
\definecolor{cvprblue}{rgb}{0.21,0.49,0.74}
\usepackage[pagebackref,breaklinks,colorlinks,citecolor=cvprblue]{hyperref}

\def\paperID{11} 
\def\confName{CVPR Workshop}
\def\confYear{2024}

\title{Detail-Enhancing Framework for Reference-Based Image Super-Resolution
}

\author{%
    Zihan Wang$^{1, 3}$, Ziliang Xiong$^{2}$, Hongying Tang$^{1}$, Xiaobing Yuan$^{2}$\\
    ${}^1$Shanghai Institution of Micro-system and Information Technology, University of Chinese Academy of Science\\
    ${}^2$Computer Vision Laboratory, Department of Electrical Engineering, LiU, Sweden \\
     ${}^3$School of Information Science and Technology, Shanghai Tech University\\
    \small{\texttt{wangzh2@shanghaitech.edu.cn, ziliang.xiong@liu.se,}}\\
    \small{\texttt{tanghy@mail.sim.ac.cn, sinowsn@mail.sim.ac.cn}}  
    }

\begin{document}
\maketitle
\begin{abstract}
Recent years have witnessed the prosperity of reference-based image super-resolution (Ref-SR). By importing the high-resolution (HR) reference images into the single image super-resolution (SISR) approach, the ill-posed nature of this long-standing field has been alleviated with the assistance of texture transferred from reference images. Although the significant improvement in quantitative and qualitative results has verified the superiority of Ref-SR methods, the presence of misalignment before texture transfer indicates room for further performance improvement. Existing methods tend to neglect the significance of details in the context of comparison, therefore not fully leveraging the information contained within low-resolution (LR) images. In this paper, we propose a Detail-Enhancing Framework (DEF) for reference-based super-resolution, which introduces the diffusion model to generate and enhance the underlying detail in LR images. If corresponding parts are present in the reference image, our method can facilitate rigorous alignment. In cases where the reference image lacks corresponding parts, it ensures a fundamental improvement while avoiding the influence of the reference image. Extensive experiments demonstrate that our proposed method achieves superior visual results while maintaining comparable numerical outcomes.

\end{abstract}





\maketitle
\thispagestyle{plain} 

\begin{figure*}[h]
      \centering
      \includegraphics[height=15cm,width=18cm]{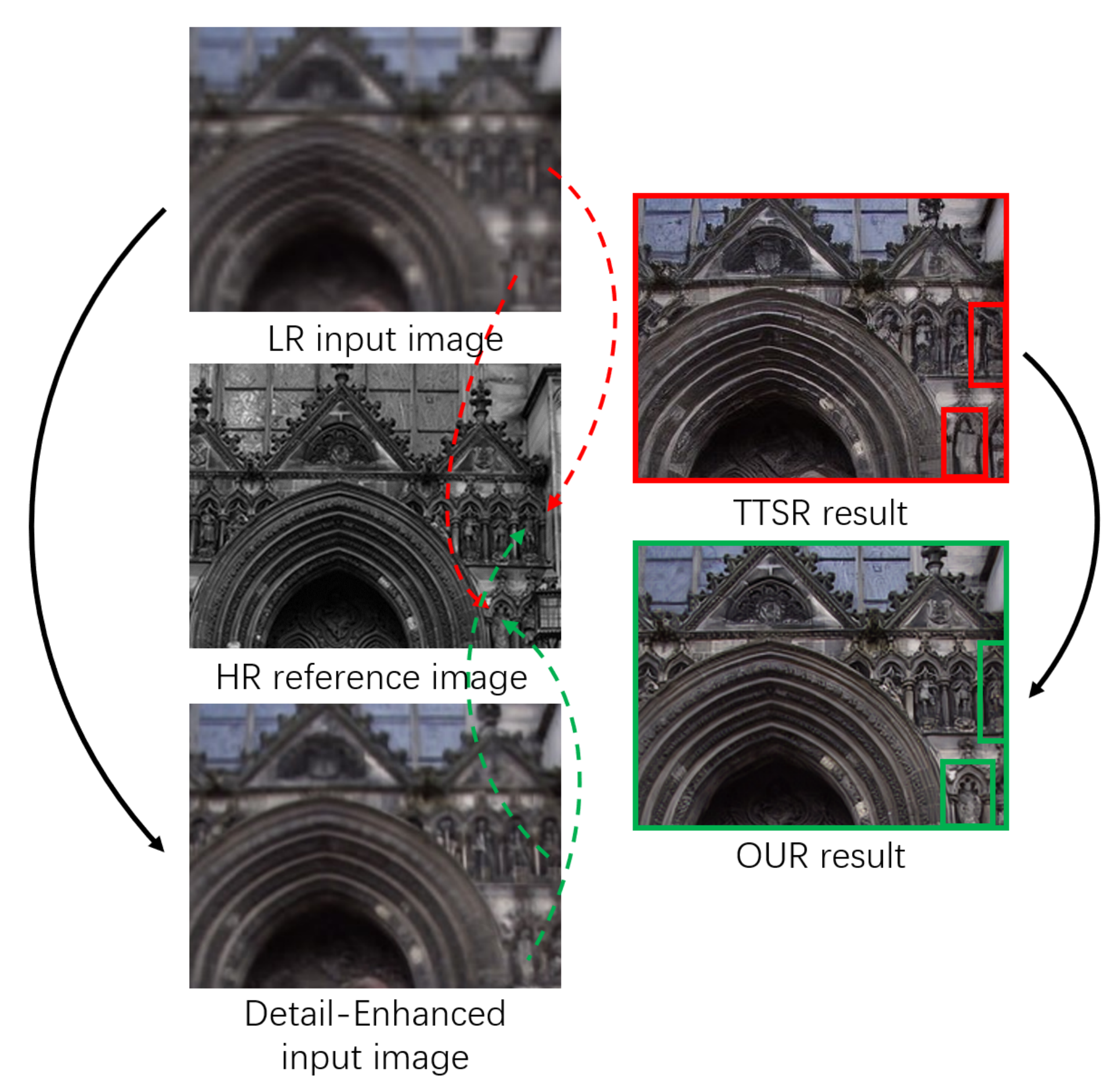}
      \caption{In the existing Ref-SR methods, such as TTSR \cite{RN21}, performance often deteriorates due to misalignment. To address this issue, we propose enhancing the fine-grained textures within images using a pre-trained diffusion model, thereby aiding the alignment process.}
      \label{fig1}
\end{figure*}

\section{Introduction}
Single image super-resolution (SISR) refers to a computational imaging technique that aims to enhance the resolution and level of detail in a single low-resolution (LR) image, typically achieved by estimating and generating a corresponding high-resolution (HR) counterpart. The essence of SISR lies in the prediction of the pixel values for the required additional pixels from the information present in a single input image. Limited by the ill-posed nature of SISR, a single LR image may generate multiple different HR images, thus resulting in the artifact and hallucination in the final output against the only corresponding ground-truth (GT) image. To ensure the authenticity of the super-resolution (SR) result, it becomes imperative to incorporate supplementary information within the reference images. The semantic information present in reference images, including content and texture, is crucial in restoring input images. Besides, obtaining similar HR reference images is significantly more feasible than acquiring strictly corresponding HR ground-truth (GT) images. To summarize, transferring HR textures from related but different HR reference images to LR input images may recover a faithful result, yielding the idea of reference-based image super-resolution (Ref-SR).

The network architecture of Ref-SR typically comprises the following four components: feature extraction, feature alignment, texture transfer, and texture aggregation. Among them, correspondence is matched between the LR image and the reference image in the feature alignment procedure, which is considered to be the most crucial component. However, image pairs consisting of the LR image and reference image do not share the same resolution, which leads to misalignment. To conquer the misalignment issue, recent advancements in this procedure have shifted the research focus of spatial alignment from point-wise matching \cite{RN19, RN27} to patch matching \cite{RN21, RN20, RN22, RN23, RN24, RN25, RN26} to improve the matching accuracy. Apart from that, in order to mitigate the resolution gap and obtain domain-consistent image pairs, previous methods tend to simply resize the input LR image to the same resolution of the corresponding reference image, e.g., bicubic interpolation. Lu \emph{et al}.\cite{RN22} choose to downsample reference images to fit into the matching process for the purpose of reducing computational complexity. While such an approach can somewhat alleviate the misalignment issue to some extent, it neglects the enhancement of details, potentially sabotaging the subsequent image-restoration results. In cases where there are corresponding features, mere resizing of LR images solely relies on the pixel values in the vicinity to predict the target pixel value, the resulting diminutive receptive field is inadequate for the comprehensive utilization of inherent information. Thus certain corresponding features between image pairs cannot be accurately aligned due to the lack of abundant details during alignment. Meanwhile, specific features within the LR image lack corresponding counterparts in the reference image, rendering them incapable of identifying aligned patches. These unaligned features remain unchanged after bicubic interpolation, hampering the visual quality of output as a result of the lack of details. Therefore, there is still room for improvement in the preprocessing of the LR image due to the absence of detail enhancement in the alignment process.

A natural idea comes up that we enhance the detail of the LR image with generative-based models ahead of alignment. Within the domain of SR, prevalent generative-based models primarily encompass two paradigms: generative adversarial networks (GANs) and diffusion models. Compared with GAN-based models, diffusion models \cite{RN17} are more stable and robust to various distributions of images. Even though diffusion models exhibit ideal performance in generating details, owing to the ill-posed nature of SR, the intrinsic randomness of diffusion models, and the deficiency in generalization capability, it is prone to generate artifacts.

To address the misalignment issue and the artifact issue altogether, we first employ theoretical analysis to elucidate the significance and positioning of details in the task of image SR. Then we propose a novel framework, which we dub, Detail-Enhancing Framework (DEF), for reference-based methods that replace the resizing of LR images with a pre-trained diffusion model. The modification of the model structure compensates for inherent limitations from both perspectives. As for Ref-based models, the introduction of diffusion models enriches the detail-wise information within the LR images, thereby benefiting the alignment between the LR image and reference image. In the meantime, to remove the artifacts, textures are transferred from reference images to guide the output of diffusion models as a procedure in the Ref-based model. Experiments have been done on five benchmark datasets, including CUFED5, Manga109, Urban100, Sun80, and WR-SR. Results demonstrate that our proposed framework attains better visual quality with comparable performance with state-of-the-art methods quantitatively.


To summarize, our primary contributions are as follows:
1)	We conduct an in-depth investigation into the significant importance of detail enhancement in Ref-SR, which has been overlooked in previous approaches.
2)	We proposed the Detail-Enhancing Framework (DEF) that introduces the diffusion models into the Ref-SR models, which not only facilitates a more precise alignment but also reduces artifacts for LR images after alignment.
3)	Experimental results demonstrate that our proposed method achieves leading visual performance while maintaining comparable numerical fidelity.

\label{}

\section{Related Work}
We will briefly review the historical issue of image SR in this section. Image SR can be roughly classified into two categories: 1) Single image super-resolution (SISR) and 2) Reference-based Image Super-Resolution (Ref-SR).


\subsection{Single Image Super-Resolution}
Restoring the information of the given LR images is the primary concern of SISR methods. 
The emergence and prosperity of deep learning contribute to the progress of SISR to a large extent. SRCNN \cite{RN1} is the pioneer of applying deep learning to SR area. Extensive research based on CNN \cite{RN2,RN3,RN4,RN5} mostly focuses on prolonging the depth of the network. 
Limited by the structure of CNN and the design of the loss function, the visual quality of the result did not improve. Then some researchers resorted to GAN-based methods \cite{RN6,RN7,RN8} and introduced perceptual loss \cite{RN9} and adversarial loss \cite{RN6}, which greatly enhance the perceptual quality. 
However, GAN-based methods require a lot of time and are hard to train, so GLEAN \cite{RN11} and PULSE \cite{RN12} utilize latent-bank to reduce consumption and formulate specific categories of images. 
In the meantime, transformer \cite{RN14} was also introduced to computer vision \cite{RN13}. To be more accurate, the implementation in the SR field includes swinSR \cite{RN15} and \cite{RN16}, etc. 
Most recently, diffusion models \cite{RN17,RN18} have proved to be efficient in generating details in the SR process, which include a forward process employed for training and a reverse process utilized for inference. 
As generative models, diffusion models \cite{RN31} share difficulty in the training process with GAN-based models, but the latter ones tend to suffer the threat of mode collapse and posterior collapse due to the additional training of discriminator, making the diffusion models the more stable ones.

\subsection{Reference-based Image Super-Resolution}
Without additional information, SISR tends to suffer the hallucination and artifacts caused by unsubstantiated prediction of pixel value. To conquer this issue, Ref-SR transfers details from relevant reference images to input images. Crossnet \cite{RN19} first proposed an end-to-end CNN network with cross-scale wrapping to achieve pixel-level alignment. However, different images may share pixels in mismatched areas which affects the construction of long-distance correlation. Thus patch-level alignment is utilized in subsequent methods. SRNTT \cite{RN20} laid the foundation of patch-level transferring which has proved to be more effective. TTSR \cite{RN21} inherited the cross-scale aggregation from SRNTT and introduced transformer and attention mechanism into Ref-SR, achieving more precise feature transfer. MASA \cite{RN22} took the potential large disparity in distributions between the LR and reference images and computation efficiency into consideration, raising coarse-to-fine correspondence matching schemes, which is also adopted by AMSA \cite{RN23}. Yet the correspondence matching still lacked robustness due to the transformation gap, so C2-matching \cite{RN24} brought in contrastive learning and knowledge distillation for better performance.

To the best of our knowledge, we find the current Ref-SR methods failed to fully exploit the detail in input LR images during the matching process. Brutely resizing the LR images only narrows the resolution gap between LR images and reference images, yet ruins the correlation between them at the detail level. Inspired by recent works \cite{RN27,RN24,RN25,RN26}, we apply Deep Convolutional Networks (DCN) \cite{RN28,RN29} in our network which requires explicit edges and contours. With the aid of our detail-enhanced input images, our approach facilitates the achievement of substantially improved alignments through the utilization of DCN. This solution effectively addresses the aforementioned challenge, offering a highly promising resolution to a significant extent.

\section{Analysis of the Super-Resolution framework}

\subsection{Range-null space decomposition}
The visual quality of the results of image SR has long been a tricky issue since it is too complex to propose a well-received metric to evaluate or improve it intentionally. Most current methods resort to narrowing the gap between pixel values, yielding an output characterized by a dearth of details, excessive smoothness, and a visual presentation that is less conducive to human perception.

Inspired by \cite{RN34,RN35}, images can be decomposed into range-space and null-space which represent the data-consistency and realness, respectively. In a more advanced context, data consistency signifies the structural characteristics of an image, while realness tends to reflect the finer details inherent in the image. Given a noise-free image SR model:
\begin{equation}
    {\bf{y}} = {\bf{Ax}}
\end{equation}

where ${\mathbf{x}} \in {\mathbb{R}^{D \times 1}}$, ${\mathbf{A}} \in {\mathbb{R}^{d \times D}}$, and ${\mathbf{y}} \in {\mathbb{R}^{d \times 1}}$ denote the ground-truth (GT) image, the linear degradation operator, and the degraded image, respectively. To derive the GT ${\mathbf{\hat x}}$ from input ${\mathbf{y}}$, two constraints have to be set to ensure the visual quality of SR:
\begin{equation}\label{eq2}
    Consistency: {\mathbf{A\hat x}} \equiv {\mathbf{y}}, Realness: {\mathbf{\hat x}} \sim q({\mathbf{x}})
\end{equation}

Where  $q({\mathbf{x}})$ denotes the distribution of the GT image.
By implementing Singular Value Decomposition (SVD) on A, we can solve its pseudo-inverse ${{\mathbf{A}}^{\mathbf{\dag }}}$ in matrix form, and the pseudo-inverse ${{\mathbf{A}}^{\mathbf{\dag }}}$ can be used to project the original image ${\bf{x}}$ to the range-space of ${\bf{A}}$ since 
\begin{equation}
    {\mathbf{A}}{{\mathbf{A}}^{\mathbf{\dag }}}{\mathbf{Ax = Ax}}
\end{equation}
                                    
conversely, ${\mathbf{(I - }}{{\mathbf{A}}^{\mathbf{\dag }}}{\mathbf{A)}}$ map ${\bf{x}}$ to null-space of ${\bf{A}}$ due to 
\begin{equation}
    {\mathbf{A(I - }}{{\mathbf{A}}^{\mathbf{\dag }}}{\mathbf{A)x = 0}}
\end{equation}

Note that any image x can be decomposed into range-space and null-space, i.e.
\begin{equation}\label{eq5}
    {\mathbf{x}} \equiv {{\mathbf{A}}^{\mathbf{\dag }}}{\mathbf{Ax}} + {\mathbf{(I - }}{{\mathbf{A}}^{\mathbf{\dag }}}{\mathbf{A)x}}
\end{equation}

\begin{figure*}[t]
      \centering
      \includegraphics[width=\textwidth]{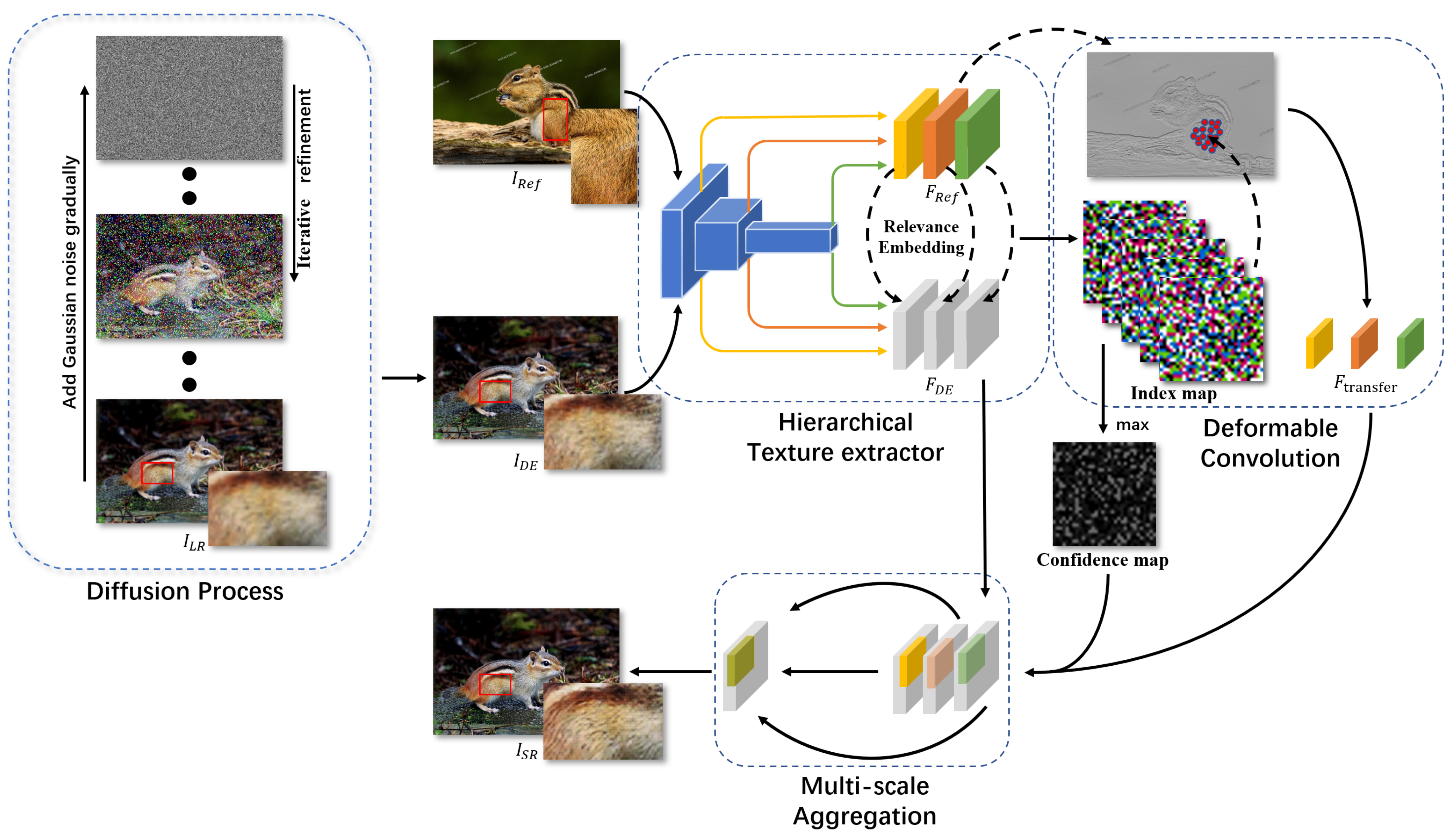}
      \caption{\textbf{Detail-Enhancing Framework overview.} For the input LR image, we commence by subjecting it to a diffusion process to enhance its fine details. Subsequently, both the detail-enhanced image and the reference image undergo feature extraction through a structurally identical network. The extracted features are aligned to obtain an index map and a confidence map, serving as the basis for the final multi-scale aggregation process.}
      \label{fig2}
\end{figure*}

\subsection{Analysis of precise details generation}
\label{section:A}
Due to the great success of PSNR-oriented models \cite{RN21,RN20, RN22, RN23}, existing Ref-SR methods tend to focus on the consistency of restored images, which is highly related to the MSE between the input and the output, rather than realness oriented from details of images. The negligence of detail generation usually leads to over-smoothed results. To investigate this issue, we retrain the TTSR\cite{RN21} model to evaluate whether details can be enhanced by the aggregation of the existing two kinds of model:

\textbf{Ref-based models}: current Ref-based algorithms can be roughly divided into three parts: feature extraction, matching, and fusion, in which fusion can be further decomposed into texture transfer and texture aggregation. By matching the most relevant patch between reference and LR images, texture can be directly transferred from HR images to LR images. This operation guarantees the data consistency of the texture that has been transferred, while there are still some drawbacks, including \textbf{texture mismatch} and \textbf{texture undermatch}.

In a reference-based dataset, textures that are similar but have slight differences may appear multiple times in the reference images. This can pose a challenge when trying to accurately match the correct texture between the reference and input images. On the other hand, reference and input images may not share the same brightness, contrast, and hue, etc. Simple execution including transferring texture without any essential adjustment can be devastating to the perceptual quality of the final output. \textbf{Texture mismatch} can occur under both circumstances.

Even under circumstances where there are multiple reference images, reference images may not cover all the textures that need to be transferred to the input images. So there could be a certain amount of unmatched textures which results in \textbf{texture undermatch}.

\textbf{Generative-based models}: Generative models and, more recently, Denoising Diffusion Probabilistic Models (DDPMs) are well-known for their ability to reproduce high-frequency details. Furthermore, in terms of reconstruction quality, it has been observed that DDPMs exhibit superior subjective perceived quality in comparison to regression-based methods \cite{RN2, RN3, RN4, RN5}, which is ideal for refining the input images in null-space iteratively. By applying general pre-trained weight, we can drastically reduce the computational complexity and acquire stable output. however, as we have mentioned above, every image has its unique distribution, which is not completely predictable by the general pre-trained weights. In this case, generative models tend to generate fake details.

To summarize, Ref-based models utilize similar reference images to guide the restoration of LR images, but the detail deficiency in correspondence matching hinders the accuracy. In contrast, the diffusion model is fully capable of generating details, whereas specific prior information is missing, resulting in artifacts in output. By aggregating the Ref-based model and diffusion model, the details generated by the latter can be utilized to enhance the correspondence map and make up for the missing details.

\section{Our Approach}

\subsection{overview}
In order to resolve the detail-enhancing issue, we propose a novel framework that inherits the main structure of Ref-based models while introducing the diffusion model. The whole detail-enhancing issue can be decomposed into two subtasks: \textbf{detail-generation} and \textbf{detail-transfer}. Intuitively, the input images are deemed to go through the reverse process of the diffusion model to form the essential details. Due to its low credibility, generated details are proportionally replaced by the corresponding part of reference images, while the rest can tackle the undermatch issue mentioned above.

For \textbf{detail-generation} task, instead of applying the downsampled input images directly which compromises the prior information severely, we upsample the input images by a pre-trained diffusion model which is known for its ability to generate rich details, obtaining detail-enhanced input images.

As for \textbf{detail-transfer} task, we follow the conventional Ref-based SR procedure. Firstly, we conduct feature extraction on both detail-enhanced images and reference images:
\begin{equation}
    {{\mathbf{F}}_{Texture}} = {F_{TE}}({{\mathbf{I}}_{{\text{Ref}}}},{{\mathbf{I}}_{DE}})
\end{equation}

where ${{\mathbf{F}}_{Texture}}$, ${F_{TE}}$, ${{\mathbf{I}}_{{\text{Ref}}}}$ and ${{\mathbf{I}}_{DE}}$ denote the texture feature, texture extraction module, reference images, and detail-enhanced images, respectively. When it comes to alignment, detail-enhanced input images are utilized to calculate the similarity between reference images and input images. The challenge associated with accurately computing the correspondence map is effectively alleviated by substituting detail-deficient input images with detail-enhanced input images. Finally, we use multi-scale aggregation module ${F_{MSA}}$ to obtain the final result ${{\mathbf{I}}_{SR}}$ from the feature of the post-transferred image ${\mathbf{F}}_{DE}$ and the feature of the reference image ${\mathbf{F}}_{Ref}$:
\begin{equation}
    {{\mathbf{I}}_{SR}} = {F_{MSA}}({{\mathbf{F}}_{Ref}},{{\mathbf{F}}_{DE}})
\end{equation}

\subsection{detail enhancement}
In contrast to the traditional Ref-SR framework, wherein feature extraction is initiated from both LR images and reference images from the outset, DEF introduces a novel paradigm by integrating a diffusion model as an initial step in the enhancement process. By decomposing the images under consideration into distinct range-space and null-space components as in (\ref{eq5}), our method prioritizes the enhancement of image details by directing primary attention toward the null-space component. We use a simple downsample operator $\mathbf{A}$ to extract the null-space information $ {\mathbf{(I - }}{{\mathbf{A}}^{\mathbf{\dag }}}{\mathbf{A)x}}$. Note that the extracted information is under the surveillance of the data consistency constraint we mentioned in (\ref{eq2}): 
\begin{equation}
    {\mathbf{A\hat x}} = {\mathbf{A}}{{\mathbf{A}}^{\mathbf{\dag }}}{\mathbf{y + A(I - }}{{\mathbf{A}}^{\mathbf{\dag }}}{\mathbf{A)}}{{\mathbf{x}}_{\mathbf{n}}} = {\mathbf{y + (A - A)}}{{\mathbf{x}}_{\mathbf{n}}} = {\mathbf{y}}
\end{equation}
where ${{\mathbf{x}}_{\mathbf{n}}}$ denotes the null-space information separated from the images. 

The operation of extracting null-space information is performed at each step during the reverse process of the diffusion model, which entails progressively recovering the image from pure Gaussian noise. At timestep t in the reverse process, the current noisy image $\mathbf{x}_t$ undergoes a denoising operation, yielding a clean image $\mathbf{x}_{{0\left|t\right.}}$. Then, a weaker noise is added to it to obtain the next noisy image $\mathbf{x}_{{t-1}}$. This iterative procedure continues until the final result $\mathbf{x}_0$ is obtained. Instead of utilizing $\mathbf{x}_{{0\left|t\right.}}$ directly, we first decompose the $\mathbf{x}_{{0\left|t\right.}}$ into null-space information ${\mathbf{(I - }}{{\mathbf{A}}^{\mathbf{\dag }}}{\mathbf{A)\mathbf{x}_{{0\left|t\right.}}}}$ and range-space information ${{\mathbf{A}}^{\mathbf{\dag }}}{\mathbf{A{x}_{{0\left|t\right.}}}}$. Then we replace the range-space information of $\mathbf{x}_{{0\left|t\right.}}$ with $\mathbf{y}$ to achieve a higher data consistency. Finally, a weaker noise is added to the combination of ${{\mathbf{A}}^{\mathbf{\dag }}}{\mathbf{y}}$ and ${\mathbf{(I - }}{{\mathbf{A}}^{\mathbf{\dag }}}{\mathbf{A)\mathbf{x}_{{0\left|t\right.}}}}$.

Subsequent to the iterative refinement of the null-space information within the diffusion model, we derive the final output by:
\begin{equation}
    {\mathbf{\hat x}} = {{\mathbf{A}}^{\mathbf{\dag }}}{\mathbf{y}} + {\mathbf{(I - }}{{\mathbf{A}}^{\mathbf{\dag }}}{\mathbf{A)}}{{\mathbf{x}}_n}
\end{equation}

Different from other generative models, the output of diffusion models encounter rigorous constraint on image size. To evaluate our method on datasets with arbitrary image size, inspired by patch-wise methodologies, we cut the images into patches that meet the requirement of image size limitations and input them into diffusion models. A logical approach that emerges involves partitioning the images into distinct patches and subsequently concatenating them during the post-processing stage. For example, if we have an image with the size of 128*256, we can cut it into two 128*128 divisions which satisfy the input demand of the diffusion model. But this will bring significant block artifacts between each division.

To conquer the question above, instead of dividing the image into unrelated patches, we take the above example and cut the 128*256 image into four 128*64 division $[{{\mathbf{y}}^{(0)}},{{\mathbf{y}}^{(1)}},{{\mathbf{y}}^{(2)}},{{\mathbf{y}}^{(3)}}]$. For the $i$-th turn we take $[{\mathbf{y}}_i^{(i)},{\mathbf{y}}_i^{(i + 1)}]$ as input and utilize diffusion model to get SR result ${\mathbf{x}}$. Then we further divide it into $[{\mathbf{x}}_i^{(i)},{\mathbf{x}}_i^{(i + 1)}]$. It is obvious that ${\mathbf{x}}_i^{(i + 1)}$ and ${\mathbf{x}}_{i + 1}^{(i + 1)}$ which represent the $i$-th and $(i+1)$-th output of same region overlap in the final concatenation, so we replace the $i$-th output with $(i+1)$-th output to ensure the coherent between each division.

\subsection{feature extraction and alignment}

To achieve precise alignment between input images ${{\mathbf{I}}_{DE}} \in {\mathbb{R}^{H \times W \times 3}}$ and reference images ${{\mathbf{I}}_{{\text{Ref}}}} \in {\mathbb{R}^{{H^{'}} \times {W^{'}} \times 3}}$, feature of both images must be extracted. By slicing the pre-trained classification model into multiple parts, we calculate the multi-scale feature of detail-enhanced images and reference images, i.e.,
\begin{equation}
    {\mathbf{F}}_{DE}^s = {F_{TE}}({{\mathbf{I}}_{DE}}),  {\mathbf{F}}_{{\text{Ref}}}^s = {F_{TE}}({{\mathbf{I}}_{{\text{Ref}}}})
\end{equation}
where ${\mathbf{F}}_{DE}^s$ and ${\mathbf{F}}_{{\text{Ref}}}^s$ are feature encoders at the s-th scale. Previous methods tend to preprocess the reference images by downsampling and then upsampling to match the frequency band. Since the diffusion model alleviates the resolution gap and reproduces rich details in the input LR images, upsampling is unnecessary.

The accuracy of alignment lies in the computation of similarity between corresponding patches. 
Cosine similarity is the most common metric for doing so. We first unfold ${\mathbf{F}}_{DE}^{s}$ and ${\mathbf{F}}_{{\text{Ref}}}^{s}$ 
into patches ${\mathbf{F}}{_{DE}^{s}}^{'} = [{q_1},...,{q_{HW}}]$ and ${\mathbf{F}}{_{{\text{Ref}}}^{s}}^{'} = [{k_1},...,{k_{H^{'}W^{'}}}]$, 
then we evaluate the relevance degree ${r_{i,j}}$ by calculating the inner product of elements in 
${\mathbf{F}}{_{DE}^s}^{'}$ and ${\mathbf{F}}{_{{\text{Ref}}}^{s}}^{'}$:
\begin{equation}
    {r_{i,j}} = \left\langle {\frac{{{{\mathbf{q}}_{\mathbf{i}}}}}{{\left\| {{{\mathbf{q}}_{\mathbf{i}}}} \right\|}},\frac{{{{\mathbf{k}}_{\mathbf{j}}}}}{{\left\| {{{\mathbf{k}}_{\mathbf{j}}}} \right\|}}} \right\rangle
\end{equation}

As for the $i$-th element in ${\mathbf{F}}{_{DE}^{s}}^{'}$, index map ${P_i}$ and confidence map ${C_i}$ can be obtained by:
\begin{equation}
    \mathop {{P_i} = \arg \max }\limits_j {r_{i,j}}, {C_i} = \mathop {\max }\limits_j {r_{i.j}}
\end{equation}

which represents the position in reference images to be transferred and the relevance degree it holds.

\subsection{texture transfer and integration}
Current Ref-SR methods encounter a pronounced decline in performance attributed to the prevalent problem of texture mismatch, as discussed in Sec.\ref{section:A}. This issue encompasses not only errors arising from the alignment procedure but also inherent deficiencies in the conventional design of convolution. Unlike regular convolution kernels, the shape of textures to be transferred may not be concrete, which leads to inaccurate mapping. To address this issue, we employ the deformable convolution network (DCN) \cite{RN24} with an adjustable receptive field. Given the position ${p_i}$ in input images, the correspondence position $p_i^k$ in index map ${P_i}$ and the confidence $c_i^k$ of transferred texture in confidence map ${C_i}$ acquired in the alignment section can be utilized to calculate $l$-th scale feature ${\mathbf{T}}_l^i$ in this position:
\begin{equation}
    {\mathbf{T}}_l^i = c_i^k\sum\nolimits_j {{w_j}} {\mathbf{F}}_{ref}^l(p_i^k + {p_c} + \Delta {p_j}){m_j}
\end{equation}

Where ${w_j}$ denotes the convolution kernel weight, ${p_c} \in \{ ( - 1,1),( - 1,0),,,,,(1,1)\}$, $\Delta {p_j}$ and ${m_j}$ denote the $j$-th learnable offset and learnable mask, respectively. After warping ${\mathbf{F}}_{{\text{Ref}}}^l$ and $l$-th scale index map ${P_l}$, $\Delta {p_j}$ and ${m_j}$ can be learned by implementing convolution on the warping result ${w_l}$ and $l$-th scale feature extracted from ${{\mathbf{I}}_{DE}}$.

Finally, the multi-scale transferred feature needs to be integrated to output the SR images. Here, we inherit the cross-scale integration module proposed by TTSR \cite{RN21} which aggregates textures from lower scale to upper scale step by step. Specifically, this module exhibits ideal performance in terms of information utilization which satisfies our claims.

\subsection{Implementation details}
The overview network can be decomposed into two sections: 1) The diffusion model which is in charge of the SISR subtask. 2) The Ref-SR architecture which includes texture extraction and transfer.

\textbf{Dataset Preprocessing.} We augment the datasets by randomly rotating images within the range of 0 to 360 degrees with intervals of 90 degrees and randomly flipping images horizontally and vertically.

\textbf{Implementation of Diffusion Model.}   We use the bicubic downsampler as the degradation operator to ensure fair comparisons. As for noise schedule and input image constraint, we choose linear noise schedule and 256*256 pre-trained model. To achieve a fine-grained diffusion process during training, we set the time step to 1,000. We avoided other time-step evaluations as they would affect comparability. 
The linear noise schedule has the endpoints of $1 - {\alpha _0} = {10^{ - 6}}$ and $1 - {\alpha _T} = {10^{ - 2}}$.


\textbf{Training of Texture Transfer Network.} For fair comparison, we train DEF on the scale of 4x, and feature extractors share the same architecture.  More specifically, we train our network using Adam optimizer with parameter ${\beta _1} = 0.9$ and ${\beta _2} = 0.999$. The learning rate is set as 1e-4, and the batch size is 9, which contains 9 LR, HR, and reference images in each batch. Note that the weight of the given extractor should be fixed, since the comparison afterward needs to be stable, and a changeable extractor can affect the performance of correspondence matching.

\textbf{Loss Function.} Given the focal point of our approach on enhancing the visual quality of the reconstructed image, coupled with the inherent emphasis on preserving intricate details through the utilization of spatial structure and semantic information of images, it becomes imperative to introduce reconstruction loss as an indispensable element, meticulously guiding the training process at its fundamental essence. To enhance the detail of SR images, perceptual loss, and adversarial loss are also introduced, so the overall loss function is written as:
\begin{equation}
    \mathcal{L} = {\mathcal{L}_{rec}} + {\lambda _1}{\mathcal{L}_{per}} + {\lambda _2}{\mathcal{L}_{adv}}
\end{equation}

To allocate greater emphasis on detail, we set weight coefficients of ${\mathcal{L}_{rec}}$, ${\mathcal{L}_{per}}$ and ${\mathcal{L}_{adv}}$ as 1, 1e-2 and 1e-4, respectively. Reconstruction loss is the only loss involved in the training process of the first two epochs for warming up the network while perceptual loss and adversarial loss are added in the following epochs to the end.
\begingroup
\setlength{\tabcolsep}{14pt} 
\renewcommand{\arraystretch}{1.2} 
\begin{table*}[]
\caption{\normalsize{Quantitative comparisons (PSNR and SSIM) of SR models}}
\label{tab1}
\begin{tabular*}{\textwidth}{c|c|c|c|c|c|c}
\hline
                        & Method     & CUFED5                                      & WR-SR                                       & Urban100                                    & Manga109                                    & Sun80                                       \\ \hline
                        & SRCNN      & 25.33/0.745                                 & 27.37/0.767                                 & 24.41/0.738                                 & 27.12/0.850                                 & 28.26/0.781                                 \\
                        & EDSR       & 25.93/0.777                                 & 28.07/0.793                                 & 25.51/0.783                                 & 28.93/0.891                                 & 28.52/0.792                                 \\
                        & ESRGAN     & 21.90/0.633                                 & 26.07/0.726                                 & 20.91/0.620                                 & 23.53/0.797                                 & 24.18/0.651                                 \\
\multirow{-4}{*}{SISR}  & RankSRGAN  & 22.31/0.635                                 & 26.15/0.719                                 & 21.47/0.624                                 & 25.04/0.803                                 & 25.60/0.667                                 \\ \hline
                        & SRNTT      & 25.61/0.764                                 & 26.53/0.745                                 & 25.09/0.774                                 & 27.54/0.862                                 & 27.59/0.756                                 \\
                        & TTSR       & 25.53/0.765                                 & 26.83/0.762                                 & 24.62/0.747                                 & 28.70/0.886                                 & 28.59/0.774                                 \\
                        & MASA       & 24.92/0.729                                 &                                             & 23.78/0.712                                 & 27.34/0.848                                 & 27.12/0.708                                 \\
                        & C2-matcing & 27.16/0.805                                 & {\color[HTML]{FD6864} \textbf{27.80/0.780}} & 25.52/0.764                                 & 29.73/0.893                                 & 29.75/0.799                                 \\
\multirow{-5}{*}{Ref-SR} & OURS       & {\color[HTML]{FD6864} \textbf{27.47/0.826}} & 27.60/0.777                                 & {\color[HTML]{FD6864} \textbf{25.92/0.780}} & {\color[HTML]{FD6864} \textbf{30.21/0.893}} & {\color[HTML]{FD6864} \textbf{29.77/0.800}} \\ \hline
\end{tabular*}
\end{table*}

\begin{figure*}[htbp]
      \centering
      \includegraphics[width=\textwidth]{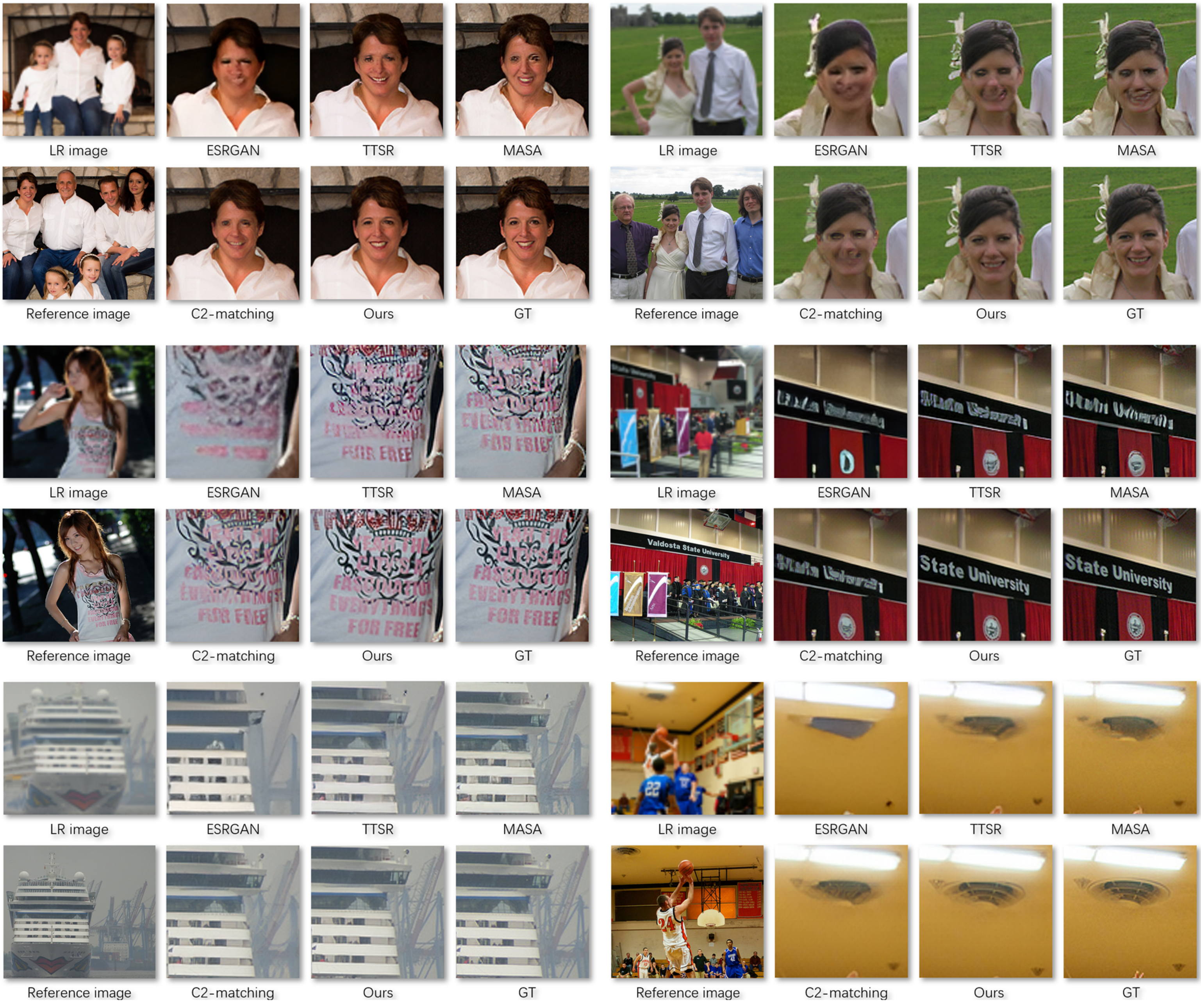}
      \caption{Visual comparison with other methods. We zoom in on the key areas for a better view.}
      \label{fig3}
\end{figure*}

\section{Experiments}
\subsection{Dataset and metrics}
\textbf{Training Dataset.} The entire training process of our model is completed on the CUFED5 \cite{RN30} dataset, which consists of 11871 pairs of images, including an input image and a reference image for each pair. Since the resolution of both input and reference images is 160x160, we resize the images in the input folder to 40x40 for the upcoming x4 SR.

\textbf{Testing Dataset.} To demonstrate the generalization ability of our network, we adopt five test sets, including CUFED5 \cite{RN30}, Sun80 \cite{RN31}, Urban100 \cite{RN32}, Manga109 \cite{RN33} and WR-SR \cite{RN24}. The test set of CUFED5 has 126 images, each image has 4 reference images with different similarity scales. Proposed by Shim \emph{et al}. \cite{RN24}, WR-SR has 80 pairs of images, each containing an input image and a reference image. The source of reference images is the most relevant search result of Google. Sun80 \cite{RN31} also has 80 natural images, each paired with multiple reference images. Urban100 has 100 building images and Manga109 has 109 manga images in which style is shared in most images. They are SISR datasets which have no reference image, thus we follow settings in \cite{RN21}: Urban100 adopts its LR images as reference, while in Manga109 we randomly select another HR image as its reference.

\textbf{Evaluation Metrics.} We evaluate the outcomes achieved by both our proposed approach and alternative methods through the application of PSNR and SSIM metrics. For a more comprehensive review of these metrics, readers could refer to \cite{RN37}. Specifically, these metrics are computed on the luminance (Y) channel of the YCrCb color space.

\subsection{Comparison with State-of-the-art Methods}
We compare our method with the previous state-of-the-art SISR methods and the single-reference Ref-SR method. SISR methods include SRCNN, EDSR, ESRGAN, and RankSRGAN, in which half of the methods we select are Gan-based due to their strong ability to generate rich details. Single-reference Ref-SR methods include SRNTT, TTSR, MASA, and C2-matching.



\textbf{Quantitative Comparison.} For a fair comparison, we train all the candidate methods on the CUFED5 dataset and evaluate them on the testsets of CUFED5, Manga109, Sun80, Urban100, and WR-SR. The scale factor of all mentioned methods is x4.
Tab. \ref{tab1} indicates that our methods outperform most of the previous state-of-the-art methods and achieve comparable performance against C2-matching on the WR-SR dataset, which emphasizes the superiority of the unique detail-generating structure we propose in the feature alignment and aggregation process. However, numerical inferiority does not necessarily imply a lack of detail. As previously mentioned, image restoration can be divided into two components: range-space and null-space. Numerical metrics primarily correspond to the range-space aspect, which is not the main focus of our proposed approach. The WR-SR dataset consists of 150 images selected from another dataset and website, serving as query images to retrieve 50 similar images from Google Images. Furthermore, these similar images undergo size normalization, resulting in 80 pairs of image sets. The normalization process undoubtedly compromises the details in the reference images, further affecting subsequent alignment procedures. Additionally, images sourced online exhibit differences in various aspects, such as lighting and contrast, making them more suitable for C2-matching due to their robustness under different conditions.

\textbf{Qualitative Evaluation.} Fig.~\ref{fig3} shows the visual results of our method, a SISR method, and previous state-of-the-art Ref-SR methods. We compare our method with ESRGAN, TTSR, MASA, and C2-matching. By comparing the selected part of the result from the same input LR image, it is obvious that our method can restore more accurate detail in various aspects. The first row of Fig.~\ref{fig3} focuses on the synthesis of natural human faces, while the focal point of the second and third rows are the recovery of letters and object textures. ESRGAN's incapacity to thoroughly exploit information from reference images results in its failure to generate reliable details. TTSR, MASA, and C2-matching can not fully utilize the information in reference images due to their detail-wise gap between input LR images and reference images, which in turn hamper the alignment and transfer procedure. For Ref-based methods, detail-enhanced input images smooth the edge of objects, which makes the alignment more accurate in the feature domain, thus reinforcing the transfer and integration procedure, exhibiting a higher visual quality image in the end.

\textbf{Alleviation to the texture mismatch and texture undermatch issues.} From the images in the first row, it can be observed that the reference images contain more faces than the images to be restored, and in the fourth image, the lighting conditions between the two images are dissimilar. In previous Ref-SR methods, this could easily lead to texture mismatch issues, ultimately resulting in distorted recovered information. However, it can be noted that through the pre-enhancement of fine details in the LR images, the accuracy of the final alignment is significantly improved, leading to more satisfactory restoration results. Additionally, in the first image, the background behind the individuals and the light tube in the last image does not have corresponding parts in the reference images to assist in the recovery of the original image. This presents a challenge related to texture undermatch. Similarly, following the enhancement of details in the LR images, regions that originally lacked corresponding HR information have also been partially reconstructed, ensuring an enhancement in the overall image restoration performance.

\subsection{Ablation study}
In this section, we conduct an ablation study to validate the effectiveness of our improvements on the baseline. Including detail-enhancing framework and feature transfer module.
\begingroup
\setlength{\tabcolsep}{8pt} 
\renewcommand{\arraystretch}{1} 
\begin{table}[]
\caption{\normalsize{Quantitative evaluation for ablation study of the decouple framework.}}
\label{tab2}
\begin{tabular}{c|c|c|c|c}

\hline
Method           & DEF & DCN & PSNR  & SSIM  \\ \hline
Base             &     &     & 25.53 & 0.765 \\
Base + DEF       & \checkmark &     & 27.37 & 0.816 \\
Base + DCN       &     & \checkmark & 27.39 & 0.819 \\
Base + DCN + DEF & \checkmark & \checkmark & 27.47 & 0.826 \\ \hline
\end{tabular}
\end{table}

\subsubsection{Detail-enhancing framework}
Instead of resizing the input LR images simplistically, our detail-enhancing framework alleviates the resolution gap by applying a diffusion model before feature extraction. We re-implement TTSR as our baseline. Ablation results are shown in Tab.~\ref{tab2}. The table reveals a substantial increase in both PSNR and SSIM values, exceeding 2dB. Previous methods usually upsample the LR image by bicubic interpolation, which exploits the surrounding 16 pixels to generate a target pixel value, matching the resolution between input images and reference images. Though the basic alignment requirement has been satisfied, the over-smooth image tends to produce artifacts in the final output. Results demonstrate that DEF outperforms the baseline by a large margin, verifying the feasibility of detail-enhancing tasks in the alignment and transfer section.

\subsubsection{Feature transfer module}
Due to the preprocess of reference images to obtain domain-consistent images between reference images and LR images, the baseline adopts transformer for alignment. To preserve the detail in reference images, we keep the original reference images, thus the transformer structure is unnecessary. We adopt relevance embedding in acquiring index map, and then according to the index, we upgrade the convolution networks to the deformable convolution networks, strengthening its robustness to irregular texture transfer. The statistic in Tab. \ref{tab2} exhibits significant improvement in performance in PSNR. Since detail has been sabotaged in the preprocessing of reference images, the enhancement of SSIM is limited.

\section{Conclusion}
In this paper, we propose a novel detail-enhancing framework to alleviate the hamper to reconstruction quality by the ill-posed nature of SR. Based on the theoretical analysis, we set two criteria in an ideal SR model to guarantee the realness and data consistency of the SR image. Specifically, we decompose the image and refine the partial content iteratively in DEF with the assistance of the diffusion model. By implementing the new framework, we are able to generate rich details in LR images and resolve the mismatch and undermatch issues in the feature alignment stage. Furthermore, the deformable convolution network is utilized to accomplish a more precise feature transfer between detail-enhanced LR images and reference images. Experiment results, especially qualitative results, demonstrate the feasibility of our proposed framework in optimizing the current Ref-SR structure.









{
    \small
    \bibliographystyle{ieeenat_fullname}
    \bibliography{main}
}


\end{document}

%% file: preamble.tex
\usepackage[dvipsnames]{xcolor}